\documentclass[conference]{IEEEtran}
\IEEEoverridecommandlockouts
\usepackage{cite}
\usepackage{hyperref} 
\usepackage{amsmath,amssymb,amsfonts}
\usepackage{algorithmic}
\usepackage{graphicx}
\usepackage{textcomp}
\usepackage{xcolor}
\usepackage{url}
\usepackage{float}
\usepackage{enumerate}
\def\BibTeX{{\rm B\kern-.05em{\sc i\kern-.025em b}\kern-.08em
    T\kern-.1667em\lower.7ex\hbox{E}\kern-.125emX}}
\begin{document}

\title{Infrared image pedestrian target detection based on Yolov3 and migration learning}

\author{\IEEEauthorblockN{Geng Shengqi}\\ Chongqing University}

\maketitle

\begin{abstract}
	With the gradual application of infrared night vision vehicle assistance system in automatic driving, the accuracy of the collected infrared images of pedestrians is gradually improved. In this paper, the migration learning method is used to apply YOLOv3 model to realize pedestrian target detection in infrared images. The target detection model YOLOv3 is migrated to the CVC infrared pedestrian data set, and Diou loss is used to replace the loss function of the original YOLO model to test different super parameters to obtain the best migration learning effect. The experimental results show that in the pedestrian detection task of CVC data set, the average accuracy (AP) of Yolov3 model reaches 96.35\%, and that of Diou-Yolov3 model is 72.14\%, but the latter has a faster convergence rate of loss curve. The effect of migration learning can be obtained by comparing the two models.
\end{abstract}

\section{Introduction}
In recent years, with the development of various industries towards intelligent direction, autonomous driving has also become the key development and transformation direction of the traditional automobile industry. In the complex driving environment, automatic driving requires extremely high accuracy and speed of pedestrian detection, otherwise it is easy to lead to traffic accident tragedy. Therefore, finding the image detection method that can effectively judge and locate pedestrian is the key task for the development of automatic driving technology.

At present, pedestrian detection is mainly based on visible light imaging data provided by vehicle-mounted optical imaging system. However, visible light imaging is greatly influenced by environmental factors. In severe weather conditions, such as heavy rain, heavy snow, thick fog, and extremely strong or weak light, visible light images will lose some details. Meanwhile, it is also limited by the need for pedestrian detection when driving at night. To solve this problem, the adoption of night pedestrian identification technology using far-infrared night vision camera~\cite{1} has significant advantages, and the processing of infrared images collected ensures the accuracy of pedestrian detection.

In this paper, I directly trained the CVC infrared data set on the basis of the Yolov3 model by using the deep transfer learning method, trained the marked CVC pedestrian data set on the basis of the pre-training weight provided, and replaced the loss function of the original model with Diou loss. Diou has the advantages of scale invariance and faster loss convergence, etc., which make up for the lack of slow convergence speed of the original loss function. Based on Diou loss function YOLOV3 transfer learning solutions of the model greatly accelerates the convergence speed and convergence degree of loss, both before and after I will replace loss function through the experiment model is applied to the infrared image at the same time the pedestrian detection, and the two models after the experiment, comparing the results of evaluation index proves that two models have their own advantages and disadvantages.

\section{Related Work}
At present, researches on infrared image pedestrian detection are mainly divided into two categories: one is based on traditional machine vision method, and the other is based on deep learning method. The application of the latter in infrared image pedestrian detection technology is gradually maturing and improving.

In the field of traditional machine vision, Ma Ye~\cite{2} proposed the background subtraction method based on the improved Gaussian mixture model (GMM) to segment human targets, and at the same time, he realized the correct detection of human targets in complex scenes by using Support vector machine (SVM). Su Xiaoqian~\cite{3} improved the three-frame difference method and adopted the three-frame difference method based on regional estimation to realize pedestrian detection in vehicle-mounted infrared images. Xu Ming~\cite{4} used the significance detection principle based on frequency domain to generate the region of interest graph, and trained the neural network to generate the pedestrian target probability graph, thus realizing pedestrian detection.

With the development of target detection algorithms in the field of deep learning, many algorithms are used to solve the problem of infrared pedestrian detection. The SE-MSSD framework based on SSD improvement was proposed by Liu Xue~\cite{5}. He used the deep separable coil product method and embedded SENet module in the network, which was better than the original SSD algorithm. Li Muskai~\cite{6} found that the target classification ability of YOLOv3 model was not perfect, so he integrated the idea of weighted demarcation of features in SENet into YOLOv3 to better describe pedestrian characteristics. The LenET-7 system proposed by Cui Shaohua~\cite{7}, which contains 3 convolutional layers, 3 pooling layers and 1 output layer, solves the problem of miscellaneous parameters of full convolutional neural network and improves the detection rate of infrared pedestrian images.

\section{Math}
\subsection{Introduction of yolov3 model}
In the field of target detection, Yolov3 has advantages over other algorithms. It was developed from Yolov2, yolov3, and improved on them, with significant performance improvements. The key points of YOLOV3 model are mainly reflected in the following three aspects: the use of the backbone network darknet53, the introduction of Residual network and the adoption of DarknetConv2D structure in Residual network.Yolov3 chose Darknet-53 as its backbone network, which contains a large number of 3 $\times$ 3 and 1 $\times$ 1 convolutional layers, eliminating the full connection layer and convolutional layer, which can well avoid this shortcoming. Each of its convolution parts uses the unique DarknetConv2D structure, and convolution is normalized before BatchNormalization, followed by negative non-zero activation function LeakyReLU.

\begin{figure}[h]
	\centering
	\includegraphics[width=0.8\linewidth]{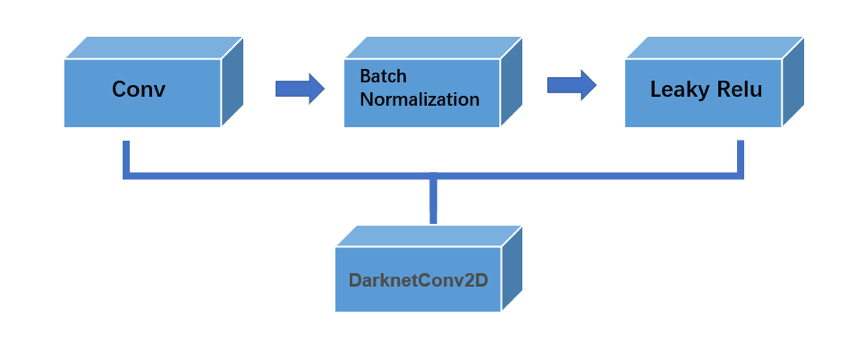}
	\caption{DarknetConv2D structure}
\end{figure}

In addition, Residual blocks in Residual network skip connection in Residual network no longer limit the increase depth of the neural network, so that the accuracy can be improved and the optimization can be improved more easily.

\begin{figure}[h]
	\centering
	\includegraphics[width=0.8\linewidth]{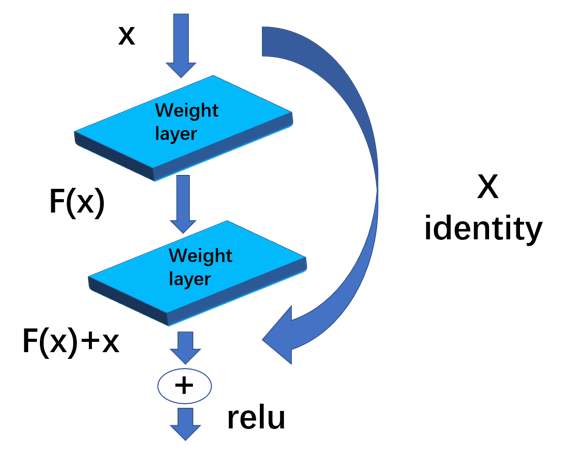}
	\caption{Residual block structure}
\end{figure}

416*416*3 image was input into the Yolov3 model and three different scales of prediction were output via the Darknet53 network. Each scale for each of the N channels contains bracanchors for each grid and each size. So we have 13 times 13 times 3 plus 26 times 26 times 3 plus 52 times 52 times 3.Each prediction corresponds to 85 dimensions, with 4 representing coordinate value, 1 representing confidence score and 80 representing coco categories. The input and output block diagram of the whole Yolov3 is as follows

\begin{figure}[h]
	\centering
	\includegraphics[width=0.8\linewidth]{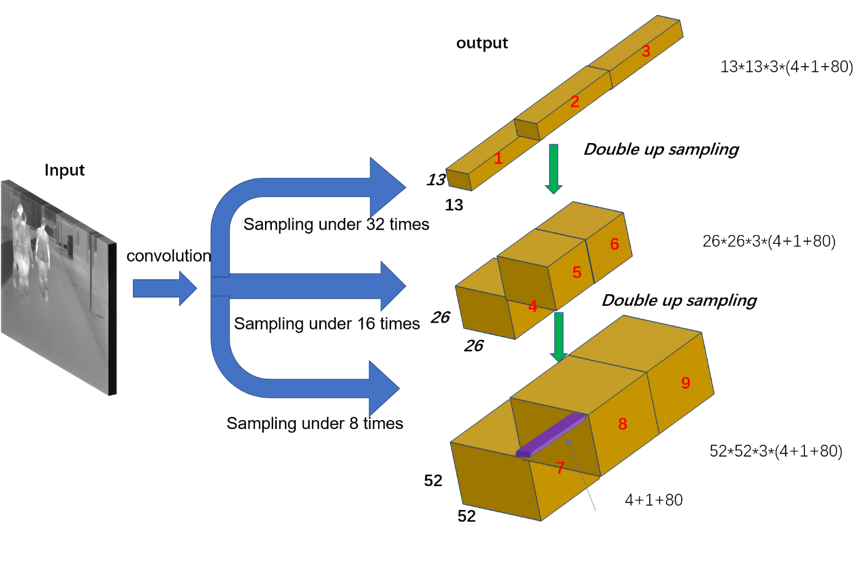}
	\caption{I/O block diagram of yolov3}
\end{figure}

\subsection{Use diou loss function}
Yolov3 adopts iou loss, which consists of three parts: the intersection ratio between the prediction frame bbox and the ground-truth is defined as iou, and then the loss function is defined as 

\begin{equation}
	L_{IoU}=1-\frac{|B \cup B^{gt}|}{|B \cap B^{gt}|}
\end{equation}

However, it can be seen from the definition that IoU loss is always 0 and IoU loss will be constant in the case that the prediction box bbox and ground-truth are non-intersecting. If the two boxes do not intersect, IoU loss is always 0. In addition, the two boxes have the same intersection area and there are many ways of intersection, so it is impossible to determine which way the two boxes intersect, which will result in the failure to determine the optimization direction. Although the Subsequent Giou proposed by Stanford scholars made up for this deficiency. But there are still many shortcomings. To solve these problems, Zheng et al. proposed Diouloss(Distance-IoU loss) in 2020~\cite{8}, and added the punishment term $R_{DIoU}$  on the basis of IOU, which is defined as follows

\begin{equation}
	R_{DIoU}=\frac{\rho^2(b,b^{gt})}{c^2}
\end{equation}

Where, the center point of $B$ and $B^{gt}$ is represented by $b$ and $b^{gt}$. $\rho(\cdot)$ is the diagonal distance. $c$ represents the minimum diagonal distance of the outer rectangleas shown in the figure 5, so diou is defined as

\begin{equation}
	L=1-IoU+R(B,B^{gt})
\end{equation}

\begin{figure}[htbp]
	\centering
	\begin{minipage}[t]{0.45\linewidth}
		\centering
		\includegraphics[width=0.9\textwidth]{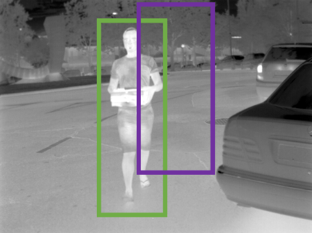}
		\caption{iou loss}
	\end{minipage}
	\hspace{20pt}%
	\begin{minipage}[t]{0.45\linewidth}
		\centering
		\includegraphics[width=0.9\textwidth]{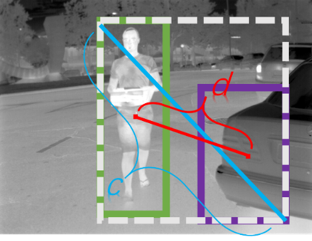}
		\caption{Diou loss}
	\end{minipage}
\end{figure}

\subsection{Transfer learning of Yolov3 model based on Diou}
The purpose of transfer learning is to apply the information learned in the source domain to the learning process of the target domain~\cite{9}.Pedestrian detection, for example, if the training from the very beginning eventually reach the purpose of infrared pedestrian detection needs a large number of data sets, high force and time loss, but if you have in large data sets, such as ImageNet migration training part of the model, model training beforehand weight was applied to the infrared image pedestrian recognition field, training on the basis of CVC infrared data sets, can quickly achieve the pedestrian detection, and the precision can be guaranteed to a certain extent. 

In this paper, the original YOLOV3 model was firstly migrated and learned. However, iou loss, the boundary box regression loss function adopted by YOLOV3, could not distinguish the optimization direction well under the condition that the intersection between the real box and the prediction box was 0 and multiple intersection were equal, and its application in infrared image pedestrian detection had some shortcomings. Therefore, aiming at the loss function problem of iou loss, this paper adopts the migration learning of The Yolov3 model based on DIOU loss. Due to the adoption of DIOU loss, the redundant prediction boxes are filtered out and the experimental results before and after are compared. It is found that the loss curve declines faster and converged faster, but the accuracy is not as good as the original model.

\section{Experiment}
\subsection{Experimental environment and data set}
The hardware and software configuration of the experimental environment is as TABLE \ref{tab-1}
\begin{table}[]
	\caption{Hardware and software configuration of the experimental environment}\label{tab-1}
	\begin{tabular}{cc}
	\hline
	\textbf{Name}             & \textbf{The related configuration}                                                                      \\ \hline
	The CPU memory            & \begin{tabular}[c]{@{}c@{}}Intel(R) Core(TM)i5-10200H CPU \\ @2.4GHz(8 CPUs), $\sim$2.4GHz\end{tabular} \\ \hline
	GPU                       & GeForce GTX 1650 Ti                                                                                     \\ \hline
	Accelerated environmental & CUDA10.1 CUDNN7.6.5                                                                                     \\ \hline
	Operating system          & Windows 10                                                                                              \\ \hline
	data processing           & Python                                                                                                  \\ \hline
	deep learning framework   & Darknet-53                                                                                              \\ \hline
	\end{tabular}
\end{table}
The purpose of this paper is to realize infrared image pedestrian target detection, so a large number of infrared image pedestrian data sets from the driving perspective are needed. This article from the Internet to download the CVC - 09: FIRSequencePedestrianDataset~\cite{10} as migration study data sets. The content is the infrared image taken by the vehicle infrared camera instrument under the driving environment. The data set annotation file is in XML format, with only a category label "Person". The label information includes the width and height of the target box and the coordinates of the center point. In this paper, 1080 images from the CVC data set are selected as the training set for the migration learning training, and 120 images are selected as the test set to detect the migration learning effect

\begin{figure}[h]
	\centering
	\includegraphics[width=0.8\linewidth]{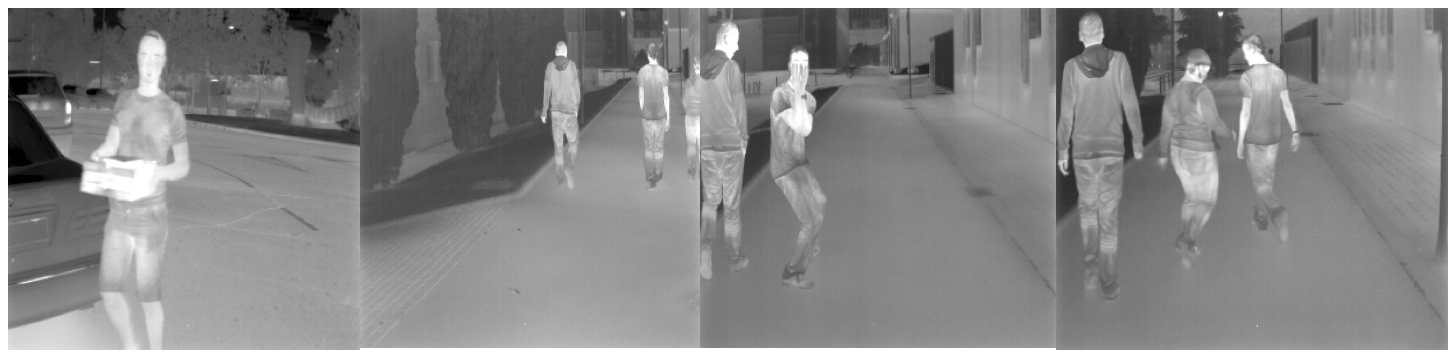}
	\caption{Part of the training set}
\end{figure}

\begin{figure}[h]
	\centering
	\includegraphics[width=0.8\linewidth]{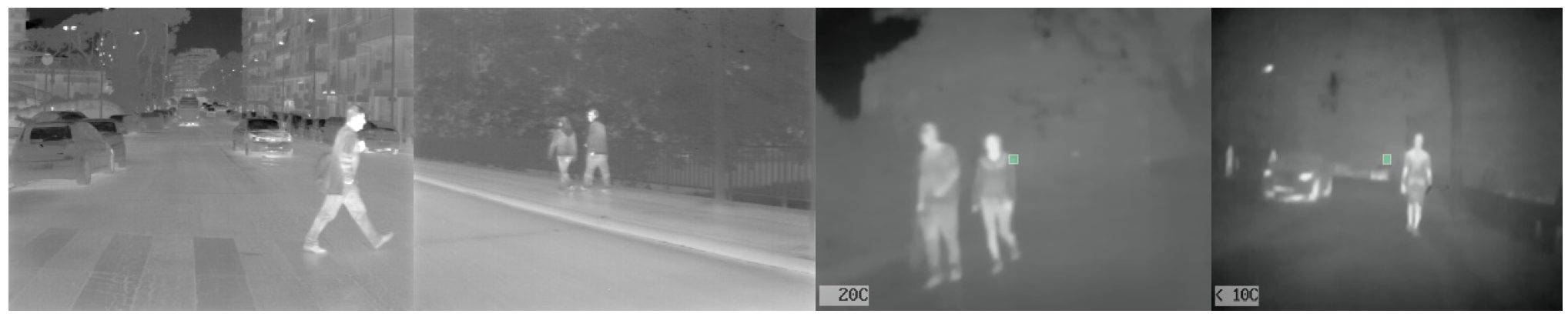}
	\caption{Partial test set sample}
\end{figure}

\subsection{Experimental evaluation index}

In order to evaluate and compare the migration learning effects of the two different models, considering that infrared pedestrian belongs to single-target detection, this paper adopts the standard evaluation index AP of single-type target detection to evaluate the detection accuracy after migration learning training. AP is defined as: taking the recall rate of infrared pedestrian as the abscissa and the accuracy rate of infrared pedestrian as the ordinate, drawing a two-dimensional coordinate curve and integrating it, and the result is AP. Accuracy is defined as 
\begin{equation}
	Precision=\frac{TP}{TP+FP}
\end{equation}
Recall rate is defined as 
\begin{equation}
	Recall=\frac{TP}{TP+FN}
\end{equation}
In Equations (4) and (5),TP represents the number of pedestrians identified on the road,FP represents the number of pedestrians identified on other objects such as vehicles and bicycles on the road, and FN represents the number of pedestrians not identified.

\subsection{Experimental process and result analysis}
This experiment mainly idea is first super parameter of the experiment was carried out, according to the results of the AP selection value using the training effect of the best parameters, different Epoch number to freeze, determine the best Freeze Epoch and Unfreeze Epoch migration study effect, after use the same training set and testing set of YOLOV3 and migration DIOU\_yolov3 model training and testing, read the same weights training files, finally compared to the average loss value curve of the two models and AP values.

\begin{figure}[h]
	\centering
	\includegraphics[width=0.8\linewidth]{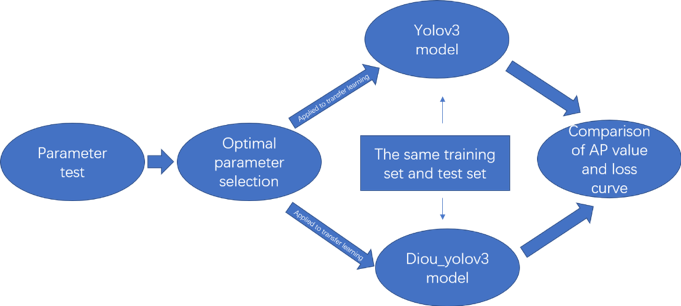}
	\caption{Experimental flow diagram}
\end{figure}

TABLE \ref{tab-2} shows the parameter test process. According to the results, in order to train the network effectively and predict the results better, the network's super parameters need to be set reasonably. The values of momentum, iteration times, batch size, and learning rate are set to optimize the hyperparameters .The set of parameters that are used to train the network are listed in Table \ref{tab-2}. Each set of values was used to train the network, and 100 images were used to test the accuracy of each result. Finally, a set of values that can maximize the quality of the network is selected as the final hyperparameter.

\begin{table}[]
	\centering
	\caption{Parameter test table}\label{tab-2}
	\begin{tabular}{cc}
	\hline
	\textbf{Parameter name} & \textbf{Parameter value} \\ \hline
	epoch                   & 25 50 100                \\ \hline
	Batch size              & 1 2 4                    \\ \hline
	learning-rate           & 0.0001 0.001 0.005       \\ \hline
	momentum                & 0.5 0.7 0.9              \\ \hline
	\end{tabular}
\end{table}
According to the accuracy of the training results of each group of parameters, epoch is 100, Freeze Epoch is 50, Unfreeze Epoch is 50, Batch size is 1, and learning-rate is 0.001 as the final parameters, which are applied to the migration learning training of the two models at the same time. The result is as Table \ref{tab-3}.

\begin{table}[]
	\caption{The value of the final selected parameter}
	\label{tab-3}
	\begin{tabular}{cccc}
	\hline
	\textbf{\begin{tabular}[c]{@{}c@{}}Epoch(Freeze Epoch \\ and Unfreeze   Epoch)\end{tabular}} & \textbf{Batch size} & \textbf{learning-rate} & \textbf{momentum} \\ \hline
	100(50+50)                                                                                 & 1                   & 0.001                  & 0.9               \\ \hline
	\end{tabular}
\end{table}

After the training, the test set was predicted in batches, and the detection results were obtained as follows. The Fig.11 shows the detection results of YOLOV3 model. It can be seen that the detection accuracy of small targets on the road is not high, there is the phenomenon of missing detection, and there is the situation of repeated detection when pedestrians overlap. The Fig.12 shows the detection results of diou-Yolov3 model. It can be seen from the detection results that, in the case of pedestrian overlap, two target boxes are well separated, which solves the problem of repeated detection.
\begin{figure}[h]
	\centering
	\includegraphics[width=0.8\linewidth]{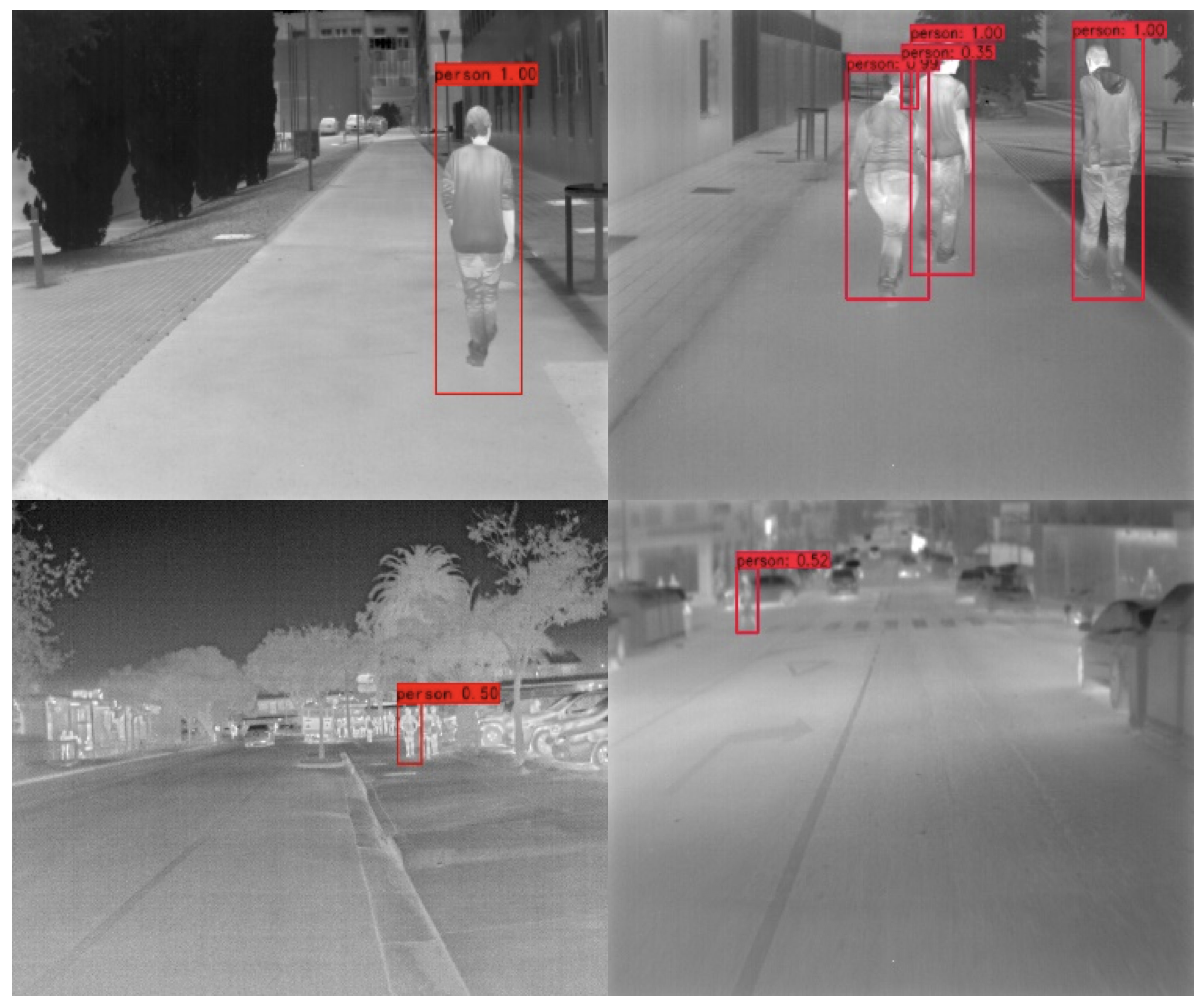}
	\caption{Yolov3 detection results}
\end{figure}

\begin{figure}[h]
	\centering
	\includegraphics[width=0.8\linewidth]{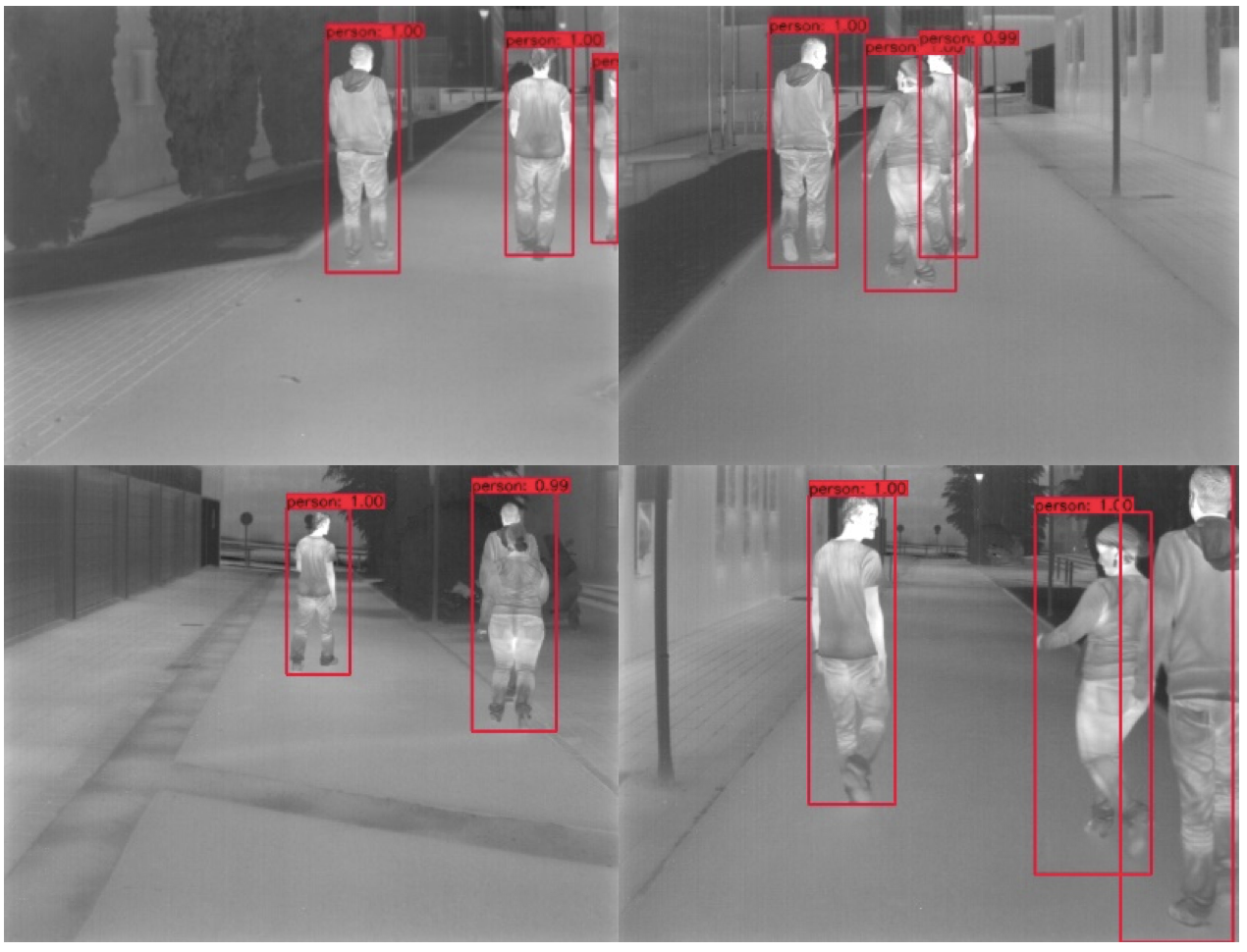}
	\caption{Diou-Yolov3 detection results}
\end{figure}

\begin{figure}[htbp]
	\centering
	\begin{minipage}[t]{0.45\linewidth}
		\centering
		\includegraphics[width=0.9\textwidth]{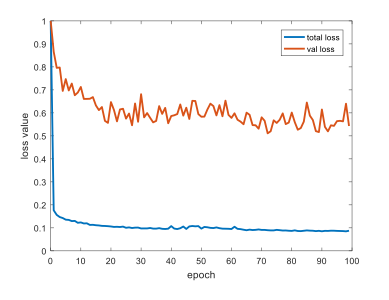}
		\caption{Yolov3 Loss curve}
	\end{minipage}
	\hspace{20pt}%
	\begin{minipage}[t]{0.45\linewidth}
		\centering
		\includegraphics[width=0.9\textwidth]{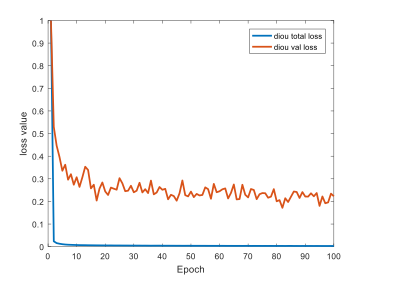}
		\caption{Diou-Yolov3 Loss curve}
	\end{minipage}
\end{figure}

In order to compare the advantages and disadvantages of the two loss functions before and after the improvement, I plotted the loss function curves of the two models on the same figure which is Fig.13. Through comparison, it can be clearly seen that compared with the loss function before the improvement, the improved DIou loss function declines faster, has better convergence and is more stable.

\begin{figure}[h]
	\centering
	\includegraphics[width=0.8\linewidth]{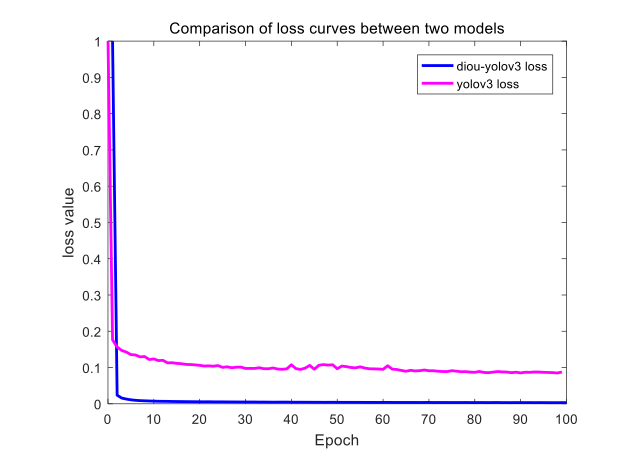}
	\caption{Comparison of loss curves between two models}
\end{figure}
Table \ref{tab-4} shows the test results of the two models. It can be seen from the table that the detection results of the two models on 120 test images are obtained, among which the AP of Yolov3 model reaches 96.35\%, and a total of 416 target pedestrians are detected, among which 347 pedestrians are correctly detected and the number of rows of error detection is 69. The AP of the Diou\_Yolov3 model reached 72.14\%, and 273 target pedestrians were detected, among which 261 rows were correctly detected and 12 were error detected. By comparison, it was found that the detection accuracy of the improved model was not as good as that of the original Yolov3 model. From the perspective of accuracy, the migration learning effect of the improved model was not ideal. 

\begin{table}[]
	\centering
	\caption{Comparison of performance indexes of two models}\label{tab-4}
	\begin{tabular}{ccccc}
	\hline
	\textbf{Network   model} & \textbf{\begin{tabular}[c]{@{}c@{}}Number of \\ predicted objects\end{tabular}} & \textbf{TP} & \textbf{FP} & \textbf{AP/\%} \\ \hline
	Yolov3                   & 416                                                                             & 347         & 69          & 96.35          \\ \hline
	Diou\_Yolov3             & 273                                                                             & 261         & 12          & 72.14          \\ \hline
	\end{tabular}
\end{table}

The following are the two model AP graphs after the program is run

\begin{figure}[H]
	\centering
	\begin{minipage}[t]{0.45\linewidth}
		\centering
		\includegraphics[width=0.9\textwidth]{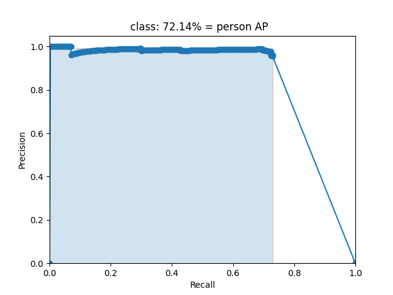}
		\caption{Diou-Yolov3 Accuracy curve of the model}
	\end{minipage}
	\hspace{20pt}%
	\begin{minipage}[t]{0.45\linewidth}
		\centering
		\includegraphics[width=0.9\textwidth]{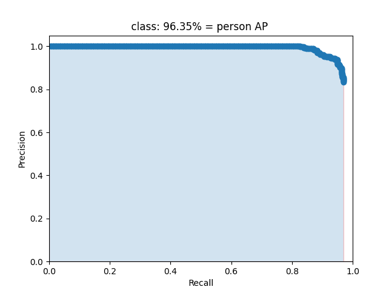}
		\caption{Yolov3 Accuracy curve of the model}
	\end{minipage}
\end{figure}

\section{Conclusion}
Based on autopilot in road pedestrian use infrared image targets detection requirements, using the two models with Diou\_yolov3 Yolov3 migration study, in has trained a large number of data set of target detection for the weights based on training, training their CVC data set, test of different parameters in the process, and the freezing and thawing different Epoch, and adopt the best parameters used in migration study training, and the freezing of different Epoch will be the result of two kinds of model training has carried on the contrast and analysis, Finally, it is concluded that the AP of the original Yolov3 model can reach 96.35\%, and the AP of the Diou\_Yolov3 model is 72.14\%. Although the loss function is improved, the detection accuracy is not high, but the curve of the loss function has a faster convergence speed, and the convergence result is lower and more stable. To sum up, the method of migration learning is used to realize the detection of pedestrians in infrared images.

\end{document}